\documentclass[journal,twoside,web]{ieeecolor}
\usepackage{jsen}
\usepackage{cite}
\usepackage{amsmath,amssymb,amsfonts}
\usepackage{algorithmic}
\usepackage{graphicx}
\usepackage{textcomp}
\usepackage{wrapfig}
\usepackage{booktabs}
\usepackage{microtype}
\usepackage{multirow}
\def\BibTeX{{\rm B\kern-.05em{\sc i\kern-.025em b}\kern-.08em
    T\kern-.1667em\lower.7ex\hbox{E}\kern-.125emX}}
\definecolor{abstractbg}{rgb}{0.89804,0.94510,0.83137}
\begin{document}
% \title{Analysis of Millimeter-Wave Frequencies on Blood Glucose Prediction Using Mixed Linear Models}
% \title{Non-Invasive Glucose Prediction Using a Mixed Linear Model and Meta-Forests: Advancements in Domain Generalization}
\title{Non-Invasive Glucose Prediction System Enhanced by Mixed Linear Models and Meta-Forests for Domain Generalization}
\author{Yuyang Sun, \IEEEmembership{Student Member, IEEE}, Panagiotis Kosmas, \IEEEmembership{Senior Member, IEEE}
\thanks{Yuyang Sun and Panagiotis Kosmas are with Department of Engineering, Faculty of Natural and Mathematical Sciences, King’s College London, London WC2R 2LS, U.K. (e-mail: yuyang.1.sun@kcl.ac.uk; panagiotis.kosmas@kcl.ac.uk).}
}
\IEEEtitleabstractindextext{%
\fcolorbox{abstractbg}{abstractbg}{%
\begin{minipage}{\textwidth}%
\begin{wrapfigure}[12]{r}{3in}%
\includegraphics[width=3in]{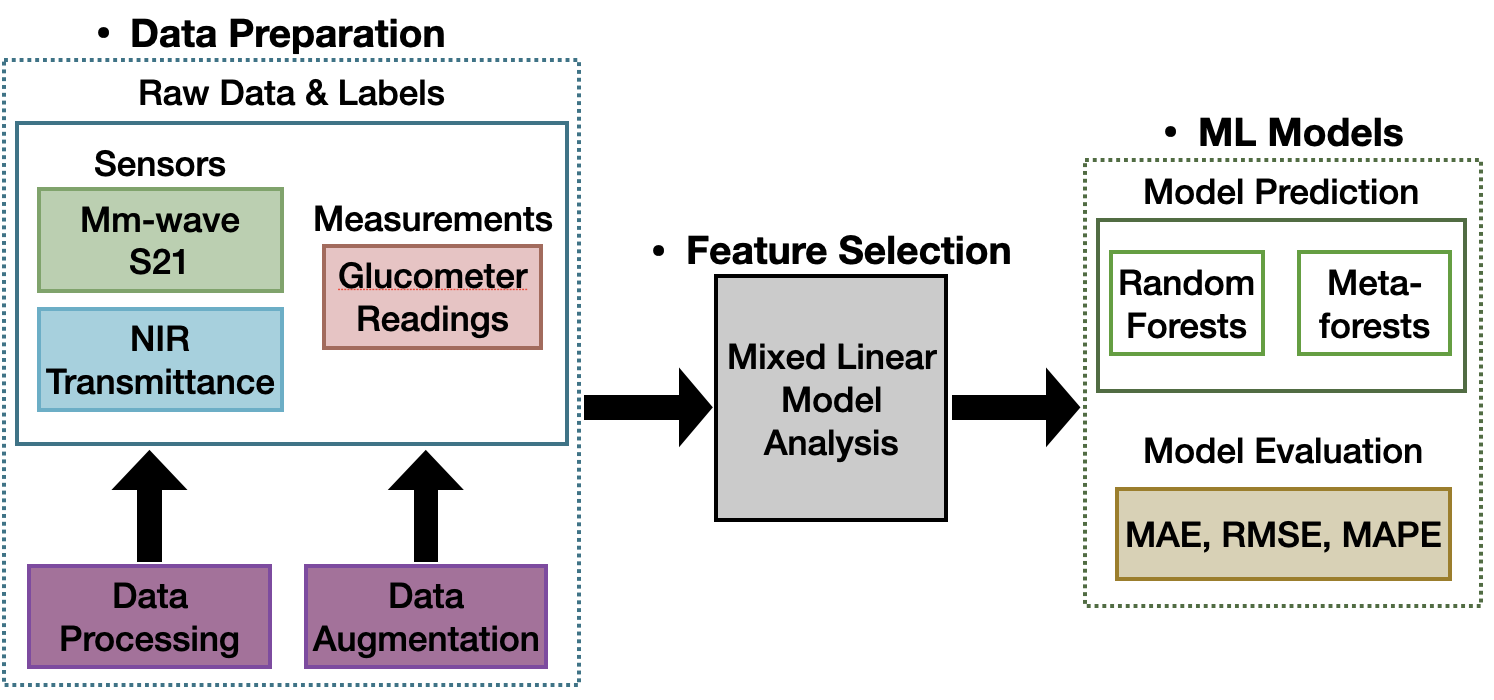}%
\end{wrapfigure}%
\begin{abstract}
In this study, we present a non-invasive glucose prediction system that integrates Near-Infrared (NIR) spectroscopy and millimeter-wave (mm-wave) sensing. We employ a Mixed Linear Model (MixedLM) to analyze the association between mm-wave frequency $S_{21}$ parameters and blood glucose levels within a heterogeneous dataset. The MixedLM method considers inter-subject variability and integrates multiple predictors, offering a more comprehensive analysis than traditional correlation analysis. Additionally, we incorporate a Domain Generalization (DG) model, Meta-forests, to effectively handle domain variance in the dataset, enhancing the model's adaptability to individual differences. Our results demonstrate promising accuracy in glucose prediction for unseen subjects, with a mean absolute error (MAE) of 17.47 mg/dL, a root mean square error (RMSE) of 31.83 mg/dL, and a mean absolute percentage error (MAPE) of 10.88\%, highlighting its potential for clinical application. This study marks a significant step towards developing accurate, personalized, and non-invasive glucose monitoring systems, contributing to improved diabetes management.
\end{abstract}

\end{minipage}}}

\maketitle

\section{Introduction}
The diabetes epidemic presents a significant global health challenge, escalating rapidly in prevalence and impacting millions worldwide \cite{cite_intro01}. Effective diabetes management relies on the ability to provide accurate, real-time glucose readings, a critical factor for timely therapeutic interventions \cite{cite_intro02}. Current glucose detection standards predominantly involve invasive methods that require frequent blood sampling throughout the day. Although effective, these approaches are marked by significant drawbacks, particularly the pain and inconvenience experienced by patients, thereby underscoring the urgent need for alternative, non-invasive glucose detection methods.

Recent advances have led to the development of various non-invasive glucose detection techniques. These non-invasive methods leverage a diverse range of biophysical and biochemical characteristics, enabling glucose level estimation without the need for blood sample extraction. These technologies include the use of optical methods \cite{cite_intro03}, photoplethysmography (PPG) signal \cite{cite_intro04,cite_intro05,cite_intro06,cite_intro07}, near-infrared (NIR) spectroscopy \cite{cite_intro08,cite_introadd01,cite_introadd02,cite_intro09}, and millimeter-wave (mm-wave) sensing \cite{cite_intro09,cite_intro10,cite_intro11,cite_intro12}, along with analyses of external fluids like sweat \cite{cite_intro13} and saliva \cite{cite_intro14}. However, methods developed for these technologies are mostly based on monitoring one or very few signals and are limited by their particular signal detection capabilities, which are affected by blood components. Moreover, the accuracy of these sensors can be affected by environmental conditions and metabolic process variations, impacting the correlation between sensor readings and actual blood glucose levels.

In comparison to single-sensor methods, multi-sensor systems provide a more robust and reliable approach to glucose level prediction by incorporating a wide range of characteristic parameters. Despite their evident potential, a notable research gap exists in the specific development of multi-sensor systems for glucose detection. By integrating multiple data types, these systems demonstrate improved robustness and accuracy, thereby addressing the inherent limitations of single-sensor systems. Among the employed technologies, NIR spectroscopy is notably prevalent in non-invasive glucose measuring. This is because there is documented evidence that NIR photons are absorbed by glucose molecules within the 900 to 1800 nm wavelength spectrum \cite{cite_intro15}. However, this wavelength range is prone to interference from blood components and water molecules, leading to signal absorption and reflection issues \cite{cite_introadd01}. To improve accuracy, mm-wave sensing can be added to NIR towards a multi-sensor system, leveraging its unique ability to discern changes in water concentration \cite{cite_intro16}. Mm-Wave technology, in particular, offers significant advantages in terms of penetration depth and is less affected by blood components, setting it apart from other non-invasive methods like NIR and PPG.

\begin{figure*}[htb!]
\centering
\includegraphics[width=0.8\textwidth,height = 4in]{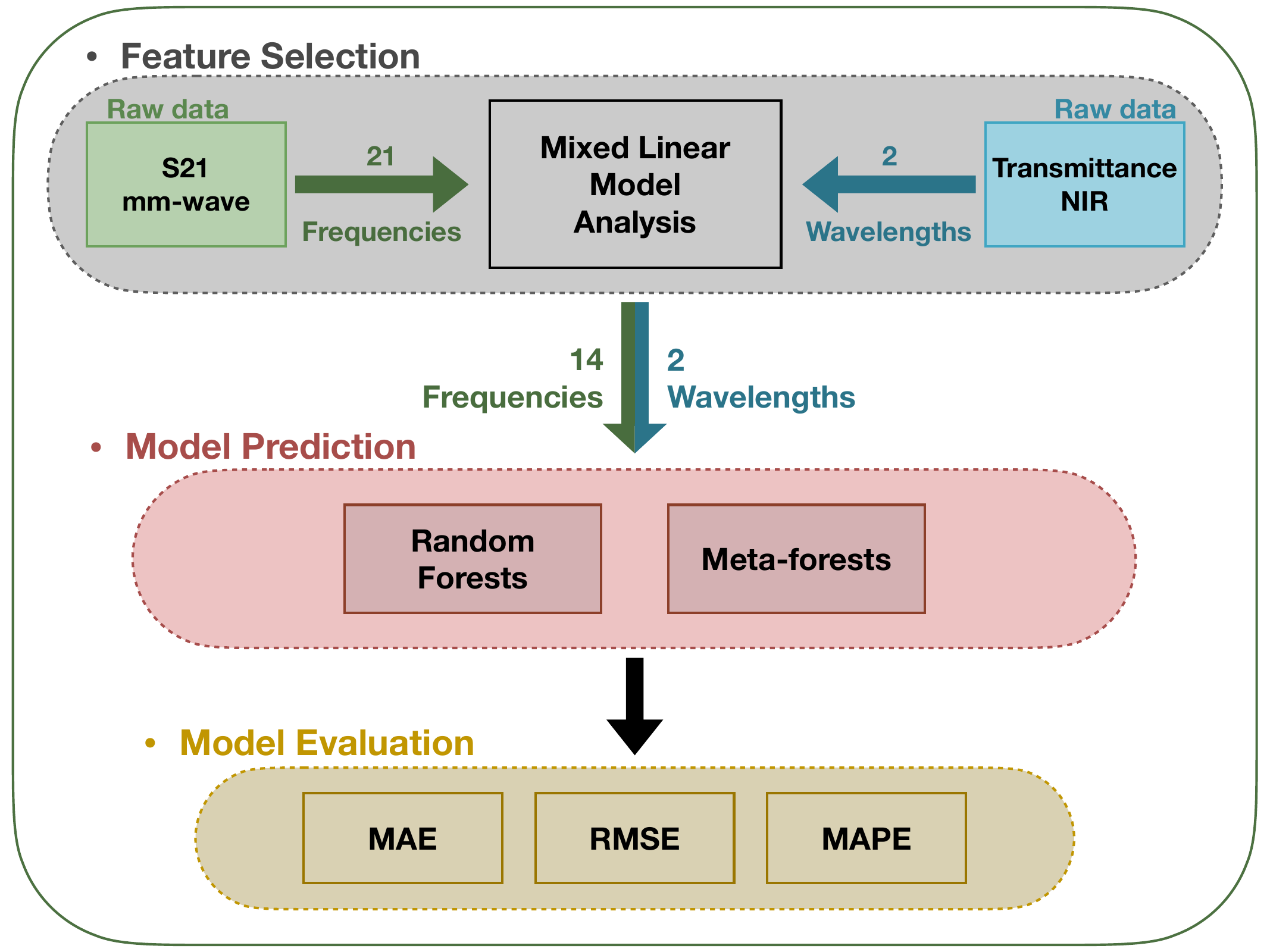}
\caption{System Overview of the Prediction Model Workflow, illustrating stages from raw data input through '$S_{21}$ mm-wave' and 'Transmittance NIR' to feature selection via Mixed Linear Model Analysis, model prediction with Random Forests and Meta-forests, and final evaluation using MAE, RMSE, and MAPE metrics.}
\label{fig:Overview}
\end{figure*}

Nevertheless, the application of mm-wave technology in glucose detection is still in a developmental phase. The primary challenge lies in the absence of definitive evidence linking mm-wave resonance with blood glucose variations and a well-defined correlation between parameters of specific mm-wave frequencies and glucose levels. The existing literature \cite{cite_intro16,cite_intro17,cite_intro18,cite_intro19} on this topic is fragmented, with a variety of frequency choices, and is lacking a comprehensive, scale-based analysis of mm-wave interactions with physiological factors.

% with auxiliary temperature and pressure sensors to mitigate the impact of ambient conditions. 

To address these challenges, our research employs a multi-sensor technique that integrates NIR spectroscopy and mm-wave sensing to provide different detection information. Employing a Mixed Linear Model (MixedLM) for an in-depth analysis of the associations between selected mm-wave frequency parameters and glucose levels, coupled with a generalized random forests model, our Meta-forests approach \cite{cite_back11} achieves low prediction errors on personalized experiments. This is evidenced by a mean absolute error (MAE) of 17.47 mg/dL, a root mean square error (RMSE) of 31.83 mg/dL, and a mean absolute percentage error (MAPE) of 10.88 \%. Figure 1 provides an overview of the workflow of our multi-sensor glucose detection system.

The proposed system shows potential in tackling the greatest challenge, arguably, in non-invasive glucose detection, which is data variability amongst different persons. The remainder of the paper is structured as follows. Section II provides background information on the ML tools used in our combined system approach, while Section III presents our methodology in data preparation, processing, and feature selection. Our main findings and results are presented in Section IV, which is followed by a discussion in Section V and a short conclusion in Section VI.

\section{Background}

\subsection{Mixed Linear Model in Statistical Analysis}
In the biomedicine area, it is common for raw datasets to be collected from different groups based on factors like subjects, populations, ethnicity, and locations \cite{cite_back01,cite_back02}. These grouping factors can result in raw data that are not truly independent. Analyzing the entire dataset without considering these grouping factors may lead to confusing results. The MixedLM extends the general linear model to incorporate both fixed and random effects, making it suitable for analyzing hierarchical and non-independent data structures \cite{cite_back03,cite_back04}. This model is widely used in various research fields to explore associations between predictor variables and response variables under different grouping factors. The MixedLM can be mathematically represented as:

\begin{equation}
Y = X\beta + Z\gamma + \epsilon
\end{equation}

Here, \( Y \) represents the response variable. \( X \) and \( Z \) are matrices of observed data associated with fixed and random effects, respectively. Coefficients \( \beta \) and \( \gamma \) denote capture fixed and random effects, respectively, and \( \epsilon \) is the error term, typically assumed to be normally distributed.

\subsubsection{Fixed Effects}
Fixed effects in MixedLM represent predictor variables expected to influence the response variable. The coefficients of fixed effects represent the estimated change in the response variable for a unit change in the predictor variable.

\subsubsection{Random Effects}
Random effects in MixedLM are grouping factors that we aim to control for. The value of random effect coefficients are not the primary focus of MixedLM. Instead, random effects are introduced for capturing the variability among different groups or subjects in the dataset, allowing the model to account for non-independence and correlation within clustered or grouped data.

For instance, Zhang et. al \cite{cite_back01} applied MixedLM to selected genetic association datasets from humans, exploring factors affecting human height by considering sex, age, and the quadratic term of age as fixed effects, with subject groups as a random effect. Similarly, Suzuki et al. \cite{cite_back02} used MixedLM on a genome-wide association study dataset including Type 2 Diabetes individuals to identify significant loci, treating gene loci as fixed effects and different ancestry groups as a random effect.

These studies demonstrate how MixedLM can be used to investigate potential impact factors strongly associated with response variables, accounting for both overall trends and individual variations. Therefore, we utilized MixedLM for feature selection in our inter-subject variability glucose dataset to identify features associated with human blood glucose changes.

\subsection{Meta-forests}
Domain Generalization (DG) aims to address the challenge of domain shift, a prevalent challenge in datasets with domain variance, particularly in biomedicine, due to individual differences \cite{cite_back05}. DG aims to minimize the model's error on unseen target domains by training on several domains with different distributions. Formally, let \(X\) denote the input space and \(Y\) the output space. We define a set of \(M\) source domains as \(D_s = \{(X_{s}^{i}, Y_{s}^{i})\}_{i=1}^M\), and a target domain as \(D_t = (X_t, Y_t)\). The objective of DG is to train a model \(f: X \rightarrow Y\) on the source domains \(D_s\), which generalizes well to the target domain \(D_t\) without requiring labeled data from \(D_t\).

The basic implementation of DG is to minimize the generalization error on unseen target domains by reducing feature dissimilarity between source and target domains \cite{cite_back06,cite_back07}. For instance, Li et al. \cite{cite_back06} introduced Maximum Mean Discrepancy (MMD) to measure and align all domain distribution distances using an adversarial learning framework. Other DG algorithms modify training strategies, such as Meta-learning \cite{cite_back08} and ensemble learning \cite{cite_back09}, to enhance generalization. Li et al. \cite{cite_back08} divided domains into multiple meta-tasks and implemented iterative meta-learning strategies to reduce domain shifts.

In contrast to these DG methods, Meta-forests \cite{cite_back10} integrate domain alignment with a modified training strategy. Specifically, Meta-forests extend the basic random forests model by incorporating the meta-learning training strategy and the MMD measure to dynamically optimize the weights of generated random forests. As established by Breiman \cite{cite_back11}, the generalization error bound of random forests is given by:

\begin{equation}
\label{eqn:1}
PE^{*} \leq \frac{\bar{\rho} \left( 1-s^{2} \right)}{s}
\end{equation}

In Equation \eqref{eqn:1}, $PE^{*}$ denotes the generalized error of random forests classifiers, 
$\bar{\rho}$ represents the mean value of correlations among classifiers in the random forests, and 
$s$ is the strength of the classifier set.

Meta-forests enhance classifier strength by optimizing the weights of generated random forests that perform well on the meta-train set. Additionally, they reduce correlations among classifiers by introducing randomness, such as feature and data sub-sampling, during the meta-train process. These mechanisms enable Meta-forests to achieve state-of-the-art accuracy with less training data compared to deep models, making it particularly advantageous for applications where data availability is limited or collection is challenging, as is the case with our glucose-collected dataset.

% Further details on Meta-forests can be found in \cite{cite_back11}.

Traditional machine learning approaches can enhance personalization by training the model individually for each subject, aiming to find an optimized fit for a specific test subject's distribution. This can be achieved by collecting labeled data from each subject over a few days, allowing the model to learn and adapt to the individual's specific glucose responses. Even with this additional personalized retraining, behavioral and physiological heterogeneities of subjects still remain. However, the Meta-forests model, a weighted collection of Random Forests, captures complex, non-linear relationships and helps to adapt to the unique glucose response patterns of each subject. 

The revised approach and the domain generalization method applied through Meta-forests aim to mitigate these heterogeneities, improving the model's robustness and accuracy. The goal of Meta-forests optimization is to find a ubiquitous distribution among all potential training sample spaces that simultaneously maintains relatively low fitting errors with the testing sample space. This objective enables the constructed Meta-forest model to handle these heterogeneities by considering various subject distributions during model training. The personalized experiment results in Table \ref{tab:my-table_3} of the revised manuscript also demonstrate the model's effectiveness in personalization. More details on the theory and proof of the Meta-forests can be found in \cite{cite_back11}.

\section{Methodology}
\subsection{Data Preparation}

% In addition to these sensing measurements, the dataset also included readings from temperature and pressure sensors. These were measured in degrees Celsius ($^{\circ}$C) and arbitrary units (a.u.) respectively, providing vital auxiliary information for the physiological state of the subjects during the tests. 

In this study, we utilized a dataset that was previously collected and detailed in \cite{cite_intro09}. The study received ethical approval from the University of Roehampton (No. LSC-2018-65). As with other related glucose monitoring datasets \cite{rodriguez2023t1diabetesgranada,zhao2023chinese}, all researchers were asked to sign a non-disclosure agreement to keep the data secure. Only selected sensing data and measured glucose readings were incorporated into this study. All subjects' personal details, such as name, age, gender, ethnicity, and medical history, were anonymized and kept private to ensure privacy and prevent any attempts to re-identify the research participants. The raw dataset comprises an extensive array of 55,273 data points, collected from two different sensor types: mm-wave and NIR sensors. These data points were obtained from five subjects (denoted as $N_{i}, i = 1, 2, ..., 5$), each participating in an intravenous glucose tolerance test, resulting in a total of 3,198 individual samples.

The mm-wave sensor data includes the $S_{21}$ transmission coefficient parameters. The $S_{21}$ parameter is associated with differential absorption rates caused by molecular rotation in water and glucose molecules. This type of dataset characterizes the transmission properties of mm-wave signals, spanning across 21 distinct frequencies, each separated by 0.25 GHz, within the range of 36.50 GHz to 41.50 GHz. For the NIR sensors, the dataset focused on the transmittance percentages at two specific selected wavelengths: 1370 nm and 1640 nm. These wavelengths are acknowledged as significant absorption points for glucose detection with less water and blood components affected. Transmittance, expressed in percentage terms, denotes the proportion of NIR optical energy penetrating through the blood sample, which is subject to both scattering and absorption processes. These features are validated by related research \cite{cite_intro15, cite_intro16} to reflect human blood glucose changes. In addition to sensor data, glucose levels were recorded as labels three times simultaneously, using both commercial (Freestyle Libre Glucose Monitoring System) and laboratory (Biosen C-line Glucose Analyzer) glucometers, with the data represented in milligrams per deciliter (mg/dL). Table \ref{tab:my-table_M1} presents the details of data and labels collected from sensors and glucometers. The sample count for glucose labels represents the number for both commercial and laboratory glucometer measurements. The 'Data Points' column indicates the number of data points collected by different frequencies, wavelengths, or devices.

% The 'Prepared Samples' column indicates the number of samples after data processing and Mix-up augmentation.

\begin{table}[htb!]
\centering
\caption{Summary of Raw Data and Labels with Details of Sample Counts. The numbers of samples listed are the aggregate of all N (N=1,2,...,5) subjects' raw samples. }
\label{tab:my-table_M1}
\begin{tabular}{cccc}
\toprule
\textbf{\multirow{2}{*}{Features} }& \textbf{\multirow{2}{*}{\begin{tabular}[c]{@{}c@{}}Collecting \\ Devices\end{tabular}}} & \textbf{\multirow{2}{*}{\begin{tabular}[c]{@{}c@{}}Raw \\ Samples\end{tabular}}} & \textbf{\multirow{2}{*}{\begin{tabular}[c]{@{}c@{}}Data\\ Points\end{tabular}} }\\
&&&    \\                 
\midrule
$S_{21}$(dB)&mm-wave &  2566  & 53886 (2566 $\times$ 21)\\
Transmittance (\%)&NIR &  509  & 1018 (509 $\times$ 2)\\
% Temperature ($^{\circ}$C)&Temperature &  2566 & 2566 (2566 $\times$ 1)\\ 
% Pressure (a.u.)&Pressure &  2566 & 2566 (2566 $\times$ 1)\\
Glucose (mg/dL)&Glucometers &  123 & 369 (123 $\times$ 3) \\            
 \bottomrule
\end{tabular}
\end{table}

\subsubsection{Data Processing}
In our research, a critical step in preparing the data from various sensors (mm-wave and NIR sensors) was the implementation of Z-score normalization. This method of normalization, also known as standardization, involved adjusting the values of each feature in our dataset so that they would have a mean of zero and a standard deviation of one. By transforming the raw sensor data onto this uniform scale, Z-score normalization effectively mitigated issues arising from disparate units and measurement ranges across different sensors. This step was crucial to ensure that each sensor's feature contributed equally to our analysis, thereby preventing any potential bias due to varying scales or variances of individual sensors.

In addition to normalization, we addressed the problem of missing values within the dataset. Missing data can significantly impact the quality of analysis, leading to biased or inaccurate results. In addressing the challenge of missing values within the dataset, we adopted a method of imputation. Missing data points were replaced with the average values from existing data for the respective sensor. This strategy was selected for its simplicity and effectiveness in maintaining the overall distribution and characteristics of the dataset. By calculating and inserting these average values, we maintained crucial trends and patterns in each sensor's data, thereby ensuring that our subsequent analysis remained robust and reflective of the underlying phenomena captured by the sensors.

Additionally, the sampling rates and collection timings of the blood glucose measurements, $S_{21}$ and NIR transmittance are different in each collection round. This variation led to inconsistencies in the time alignment among $S_{21}$, NIR transmittance, and blood glucose measurements. To address this and ensure accurate correspondences between the extracted features and labels, we adopted a delete-and-insert approach according to the size of the NIR transmittance samples.

Firstly, addressing the mismatch in sampling rates, we preserved only the nearest $S_{21}$ samples from the mm-wave sensors that corresponded temporally with the NIR transmittance samples, while discarding others. This step ensured that each NIR transmittance sample was matched with the closest $S_{21}$ parameters in terms of collection time.

Subsequently, we tackled the lower sampling rate of the glucose labels compared to the NIR transmittance samples. To address this, linear interpolation was introduced as a method to estimate glucose values based on their temporal proximity. The linear interpolation process, detailed in Formula 1, allowed for the calculation of glucose values at time points where direct measurements were unavailable.

\begin{equation}
G_j = G_i + (T_{j} - T_i) * \frac{\left |G_{i+1}-G_i  \right |}{T_{i+1}-T_i}
\end{equation}

Here, $G_j$ represents the estimated glucose level at a specific time point $T_j$. To calculate this, the formula interpolates between two known glucose measurements, $G_i$ and $G_{i+1}$, taken at two adjacent time $T_i$ and $T_{i+1}$, respectively.

After these data processing steps, the sample sizes across $S_{21}$, transmittance, and glucose readings, standardizing them to a uniform count of 509 samples from 5 different subjects.

\subsubsection{Mix-up Augmentation}
To address the challenge of unbalanced data distribution in our study, where data collected from different subjects varied significantly in volume, we employed the Mix-up method for data augmentation, as described in \cite{cite_M1}. This technique was crucial for mitigating potential model bias, especially considering that some subjects contributed a large volume of data compared to others.

Our objective with the Mix-up augmentation was to ensure the number of samples from each subject to match the count ($N_2 = 112$) of the subject with the largest dataset, post-linear interpolation. We adopted an $\alpha$ value of 0.4 for the Mix-up process, following the standard Mix-up formula:

% $x' = \lambda x_i + (1 - \lambda) x_j$ and $y' = \lambda y_i + (1 - \lambda) y_j$, where $\lambda$ is drawn from a Beta distribution Beta($\alpha$, $\alpha$). 

% enabling effective dataset augmentation and enhancing model robustness.

\begin{equation}
    x' = \lambda x_i + (1 - \lambda) x_j
\end{equation}
\begin{equation}
    y' = \lambda y_i + (1 - \lambda) y_j
\end{equation}
where \( x_i, x_j \) are feature vectors, \( y_i, y_j \) are their corresponding labels, and \( \lambda \) is a mixing coefficient drawn from a Beta distribution, Beta(\(\alpha\), \(\alpha\)). This method creates new samples \( x' \) and labels \( y' \) by linearly combining pairs of existing data.

This approach enabled the creation of synthetic data points, effectively augmenting the dataset for subjects with fewer samples and achieving a balanced representation across all subjects. As a result, we achieved a more uniform and diverse dataset across all subjects. This augmented dataset enhance the robustness and generalizability of machine learning model, especially in leave-one domain experiments. Following the Mix-up augmentation, each subject's sample count was standardized to 112 ($N_{i} = 112, i = 1, 2, ..., 5$), in a total sample size of 560 after the whole data preparation process.

% Mixup
% For label processing.

% Due to collecting frequencies of data and labels being different, the size of data and labels are different. Besides, 

% we applied the Mixup method to augment the label and align it with the collected data.

% Besides, 

\subsection{Mixed Linear Model for Feature Selection}
The complexity of the raw dataset in our study required a robust statistical approach for feature selection. We employed a MixedLM as a tool for feature selection, given its effectiveness in handling the high dimensionality of our data and the inherent individual variability among subjects. Recognizing the absence of a resonant absorption signature for glucose in the mm-wave band, the mm-wave signal attenuation is influenced by the molecular rotation of water and glucose molecules. Therefore, we focused on investigating the relationships between $S_{21}$ parameters (selected among 21 mm-wave frequencies) and blood glucose levels through MixedLM analysis.

In the implementation of the MixedLM, we designated the $S_{21}$ parameters as fixed effects, while incorporating subjects as random effects. This setting was crucial to model individual variability and to address the non-independence of measurements within the same subject. Our feature selection criteria were stringent: we only considered $S_{21}$ parameters of mm-wave frequencies with an absolute P-value less than 0.05, to ensure the robustness of our findings. The threshold $p<0.05$ was chosen because it represents a widely accepted standard for ensuring statistical significance \cite{cite_000}.

\begin{figure}[htb!]
\centering
\includegraphics[width=.48\textwidth]{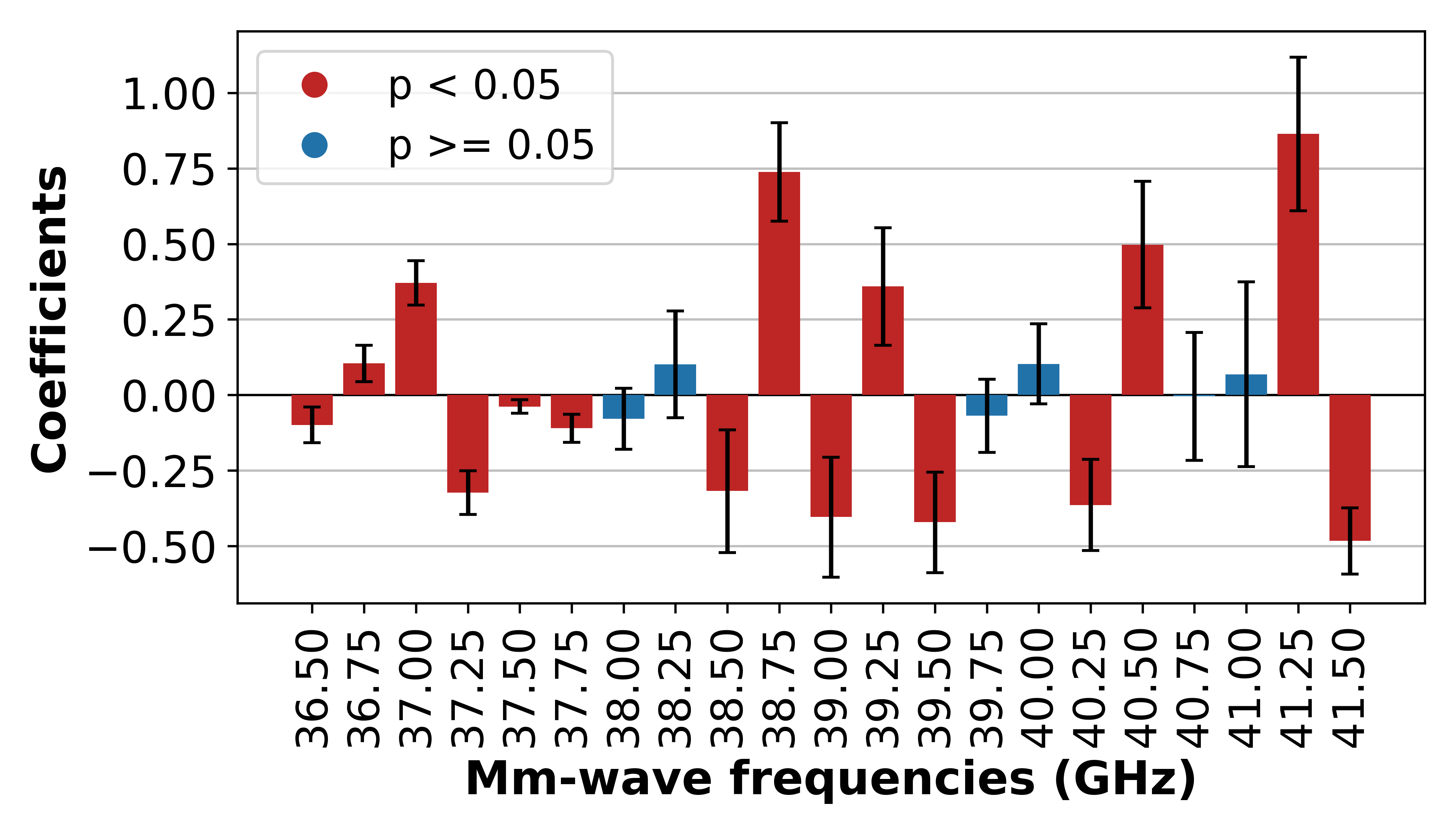}
\caption{Estimated Coefficients of $S_{21}$ Parameters for mm-wave Frequencies from Mixed Linear Model Analysis. The lower and upper error bars delineate the 95\% confidence intervals (2.5\%, 97.5\%) of estimated coefficients. The coefficient for each frequency reflects the estimated change in the response variable (normalized glucose level) for a one-unit increase in the predictor variable ($S_{21}$ parameter) while holding other variables constant. Red bars indicate frequencies with statistically significant associations (\( p < 0.05 \)), selected for further study, while blue bars represent non-significant frequencies (\( p \geq 0.05 \)), which were removed from subsequent analysis. }
\label{fig:MixedLM}
\end{figure}

The MixedLM results revealed several mm-wave frequencies meeting this criterion, which were selected. The results of the MixedLM model selection were visualized in Figure \ref{fig:MixedLM}, where red bars indicate frequencies that met our significance threshold, signifying their statistically significant associations with glucose level changes. Additionally, we applied a similar analysis to the NIR signal wavelengths. Both transmittances from the selected NIR wavelengths displayed P-values below 0.05, indicating a statistically significant effect within the NIR wavelength range, corroborating related research findings \cite{cite_intro15}.

% It's crucial to note that the statistically significant associations identified with mm-wave frequencies do not imply a direct correlation or causative relationship with blood glucose changes. Instead, these findings serve as an indication that certain frequencies in our dataset exhibited statistical associations meriting further exploration. Therefore, we selected a feature subset of mm-wave frequencies, which satisfied the P-value criteria and integrated with other NIR transmittance features for subsequent model prediction in our study.

It is crucial to note that the statistically significant associations identified with mm-wave frequencies do not imply a direct correlation or causative relationship with blood glucose changes. Instead, these findings suggest that certain frequencies in our dataset exhibited statistical associations warranting further exploration. Therefore, we used the MixedLM to analyze the features and identify a subset of mm-wave frequencies that satisfied the P-value criteria. These selected features were then integrated with other NIR transmittance features for subsequent model prediction using both Meta-forests and basic Random Forests models. The MixedLM ensured that the most associated features were chosen during the feature selection process. These features were then used to train the non-linear Meta-forests and Random Forests models, enabling them to capture complex patterns between features and labels. This integration allowed for generalizable and accurate prediction results while providing interpretability through the feature selection process.

\subsection{Models Prediction}

{\color{black}For predictive modeling, we utilized a subset of selected features to estimate glucose values at the corresponding time. The models we employed for prediction were Random Forests and an advanced meta-learning strategy known as Meta-forests. Compared to other machine learning methods, such as neural networks, tree-based models present a lower risk of over-fitting. Additionally, they offer tunable hyperparameters like 'max\_depth' and 'n\_estimators', which enable the optimization of model complexity. These characteristics are suitable to our dataset, which features a limited number of features and data points.

Both the Random Forests and Meta-forests models were implemented based on the Scikit-Learn package \cite{cite_M3}. For both models, we set the hyperparameters 'n\_estimators' to 100 and 'max\_depth' to 10. The training and testing dataset split for the Random Forests model followed a traditional 7:3 split. The Meta-forests model, designed to enhance adaptability through domain generalization, involved a further subdivision of the source domain into meta-train and meta-test sets for each meta-task iteration. Specifically, one subject domain was randomly selected as the meta-test set, while the rest source domains were served as the meta-train set. During the meta-learning training strategy, 30\% of the data from both the meta-train and meta-test sets was selected for each meta-task iteration. After several rounds of meta-task optimization, the resulting Meta-forests model underwent testing on the final target domain dataset, which remained unseen in previous training iterations.

The following experiments and results section provides a detailed account of the results derived from conducted experiments using both Random Forests and Meta-forests models. To assess the performance of these models, we introduced three key metrics: Mean Absolute Error (MAE), Root Mean Square Error (RMSE), and Mean Absolute Percentage Error (MAPE). Both MAE and RMSE are uniformed in milligrams per deciliter (mg/dL), providing a direct measure of prediction accuracy in the same units as blood glucose measurements. MAPE, expressed as a percentage, serves as a metric for evaluating the performance of glucose analyzers by quantifying the relative deviation between predicted measurements and the actual blood glucose concentrations \cite{cite_M2}, calculated as in Formula belows:}

{\color{black}\begin{equation}
    MAPE=\frac{1}{n}\sum_{i=1}^{n}{\frac{\left | x_i - y_i \right |}{y_i}} \times 100\% 
\end{equation}}

{\color{black}Here, $x$ and $y$ represent n-dimensional vectors corresponding to the reference and predicted glucose values, respectively, while $n$ denotes the size of the test sample.
}

\section{Experiments and Results}

\subsection{Experiment Setup}

In this section, we detail the experimental setup and provide an analysis of the results from two distinct series of experiments. These experiments were designed to assess the predictive performance of the proposed system and to evaluate the impact of specific mm-wave frequency selections on blood glucose prediction performance, offering valuable insights for future research directions. All experiments were conducted on the online platform Google Colaboratory to facilitate easy deployment across multiple environments. To reduce high computing resource requirements, we utilized an Intel Xeon CPU 2.20GHz instead of the NVIDIA T4 GPU option.

The first series of experiments, referred to as 'generalized model experiments', leveraged the entire dataset from all five subjects, split in a 7:3 ratio to form the training and testing sets. These experiments exclusively employed the Random Forests model. The objective was to evaluate the model's performance across the entire dataset, disregarding individual variances. The Random Forests model required an average of 0.93 seconds for training and 0.01 seconds for testing in each generalized model experiment.

The second series of experiments, termed 'personalized model experiments', adopted a leave-one-domain-out strategy \cite{cite_E1}. These experiments utilized both Random Forests and Meta-forests models. The models were trained on data from four subjects and tested on the remaining fifth subject. This procedure was repeatedly executed five times, with each subject in turn serving as the testing set, thereby enabling an evaluation of subject-specific variability. Notably, in the model training phase of each iteration, the test subject's (target domain's) data remained unseen to ensure the reliability of the experiment results. On average, the training process for each personalized model experiment with the Random Forests model took 0.78 seconds, while the testing process took 0.01 seconds. For the Meta-forests model, the average training time was 1.57 seconds, and the testing time was 0.02 seconds. The training and testing times for the Random Forests models did not vary significantly between the two series of experiments. In contrast, the training time for the Meta-forests model increased due to the additional weight matrix calculations and the larger number of trees generated compared to the basic Random Forests model.

% In this section, we detail the experiment setup and provide an analysis of the results from two distinct experimental series. These experiments were designed to assess the predictive performance of the proposed system and to show the impact of specific mm-wave frequency selections on blood glucose prediction performance, offering valuable insights for future research directions.

% The first experimental series, referred to as 'generalized model experiments', leveraged the entire dataset from all five subjects, splitting it in a 7:3 ratio to form the training and testing sets. These experiments were exclusively conducted using the Random Forests model. The objective was to evaluate the model's performance across whole datasets from various subjects by disregarding individual variances.

% The second series, termed 'personalized model experiments', adopted a leave-one-domain-out strategy \cite{cite_E1}. These experiments were conducted using both Random Forests and Meta-forests models. The models were trained on data from four subjects and then tested on the remaining fifth subject. This procedure was repeatedly executed five times, with each subject in turn serving as the testing set, thereby enabling an evaluation of subject-specific variability. Notably, in the model training phase of each iteration, the test subject's (target domain's) data remained unseen to ensure the reliability of the experiment results.

\begin{table}[htb!]
\centering
\caption{Summary of Experimental configurations, categorizing experiments into 'Generalized' and 'Personalized' series based on considering individual variance in the dataset.}
\label{tab:my-table_0}
\begin{tabular}{cccc}
\toprule
Number & Series Type & Features & Model \\
\midrule
                 1 & Generalized & All frequencies & Random forests \\
                 2 & Generalized & Selected frequencies & Random forests \\
                 3 & Generalized & Removed frequencies & Random forests \\ 
                 4 & Personalized & All frequencies & Random forests \\
                 5 & Personalized & Selected frequencies & Random forests \\
                 6 & Personalized & Removed frequencies & Random forests \\
                 7 & Personalized & All frequencies & Meta-forests \\
                 8 & Personalized & Selected frequencies & Meta-forests \\
                 9 & Personalized & Removed frequencies & Meta-forests \\
\bottomrule
\end{tabular}
\end{table}

In addition, both series were conducted using three distinct mm-wave feature sets in conjunction with NIR transmittance: 1) the complete set of mm-wave frequency features and NIR transmittance at two wavelengths, 2) a subset of selected frequency features with a significance level of \( P < 0.05 \) and NIR transmittance, and 3) the subset of non-significant removed frequency features (\( P \geq 0.05 \)) and NIR transmittance. These different feature setups were employed to directly assess the influence of associated and non-associated features on model prediction performance. For reference, Table \ref{tab:my-table_0} summarizes the detailed configurations of the experiments, including the type of experimental series, input features, applied machine learning model, and the corresponding experiment number. In Table \ref{tab:my-table_0}, the 'features' column indicates the condition of the mm-wave frequency range used, with NIR wavelengths' transmittance included in all experiments. The 'model' column specifies the forest model employed for prediction.

% \caption{Summary of Experimental configurations, categorizing experiments into 'Generalized' and 'Personalized' series based on the consideration of individual variance in the dataset. The 'features' column indicates the condition of the mm-wave frequency range used, with NIR wavelengths' transmittance included in all experiments. 'Model' specifies the forest model employed for prediction.}

% To evaluate the experiment results, we introduced three metrics, Mean Absolute Error (MAE), Root Mean Square Error (RMSE), and Mean Absolute Percentage Error (MAPE). The units of MAE and RMSE are uniformed as mg/dL. The MAPE delineated as a percentage, is a metric to evaluate the performance of glucose analyzers by calculating the relative deviation between measurements and the true blood glucose concentration \cite{b4_a1}, calculated as in Formula (1).

% \begin{equation}
%     MAPE=\frac{1}{n}\sum_{i=1}^{n}{\frac{\left | x_i - y_i \right |}{y_i}} \times 100\% \tag{1}
% \end{equation}

% Here, $x$ and $y$ represent n-dimensional vectors corresponding to the reference and predicted glucose values, respectively, while $n$ denotes the size of the test sample.

\subsection{Generalized Model Experiments Results}
{\color{black}The first series of experiments provided insights into the forecasting model's generalized capability when integrated with the frequency selection method. For this series, data from all five subjects were combined into a single dataset, and subsequently divided into training and testing sets with a 7:3 ratio. To ensure the reliability and robustness of the findings, each experiment was conducted 10 times, each with uniquely split training and testing sets. The results, presented in Table \ref{tab:my-table_1}, indicate the model's average performance across these experiments, without considering individual variances. The 'Number' column in Table \ref{tab:my-table_1} corresponds to the experiment identifiers listed in Table \ref{tab:my-table_0}.}

\begin{table}[htb!]
\centering
\caption{Summary of Results for Generalized Experiment Type.}
\label{tab:my-table_1}
\begin{tabular}{cccc}
\toprule
  Number &  MAE (mg/dL) &  RMSE (mg/dL) &  MAPE (\%) \\
\midrule
                 1 &        18.35 &          35.06&    12.63 \\
                 2 &        18.11 &         34.01 &    12.59 \\
                 3 &        24.11 &         47.45 &    15.68 \\ 
\bottomrule
\end{tabular}
\end{table}

{\color{black}Analysis of the generalized model revealed a marginal improvement in prediction accuracy upon implementing the Mixed Linear Model's P-value criterion for frequency selection. This improvement is quantified by a reduction of 0.24 mg/dL in MAE, 1.05 mg/dL in RMSE, and 0.04\% in MAPE, as shown in Experiment 2 compared to Experiment 1. Notably, the exclusion of statistically associated mm-wave frequencies resulted in a significant decline in model prediction performance. As shown by the results of Experiment 3, all evaluation error metrics increased rapidly relative to the other two generalized model experiments, emphasizing the importance of the selected frequencies in the prediction model. Therefore, the generalized model experiments suggest that the features of selected frequencies with ($P < 0.05$) play a critical role in precise prediction, while other features of frequencies with ($P \geq 0.05$) are redundant and do not significantly influence the model's predictive capacity. The following subsection provides a detailed analysis of how these models perform when applied to different individuals.}

\subsection{Personalized Model Experiments Results}

{\color{black}The personalized model experiments are designed to evaluate model performance while considering subject variance, simulating clinical situations where the test subject's data was not previously collected and trained. The details of the personalized model experiments with Random Forests and Meta-forests are as follows:

In experiments 4, 5, and 6, employing the Random Forests model, the leave-one-domain-out strategy is applied, differing from the traditional train-test split with a 7:3 ratio used in the generalized experiments. This strategy, commonly used in evaluating DG problems, involves using all data from one $N_{i}$ ($N_{i}, i = 1, 2, ..., 5$) as the testing set (target domain), while the remaining four $N_{i}$ sets constitute the training set (source domain). All the training set is utilized for model training, and the testing set evaluates the prediction model's performance.}

\begin{table}[htb!]
\centering
\caption{Ablation Study Results for Different Portions (\(P\)) of Meta-Train Data in One Meta-Task Iteration. These results are obtained under the experiment setup same as Experiment 8.}
\begin{tabular}{cccccc}
\toprule
\multirow{2}{*}{Metrics}          & \multicolumn{5}{c}{Portion ($P$)} \\ 
\cmidrule{2-6}
& $10\%$ & $20\%$ & $30\%$ & $40\%$ & $50\%$\\ \midrule
MAE            & 19.29 & 18.94 & 17.54 & 22.80 & 26.31 \\
RMSE & 35.21 & 34.23 & 32.01 & 41.61 & 48.01 \\
MAPE      & 11.63 & 11.17 & 10.57 & 13.74 & 15.86 \\
\bottomrule         
\end{tabular}
\label{tab:my-table_4}
\end{table}

\begin{table*}[htb!]
\centering
\caption{Summary of Personalized Model Experiment Results. }
\begin{tabular}{ccccccccc}
\toprule
\multirow{2}{*}{Number} & \multirow{2}{*}{Metrics}&\multirow{2}{*}{\begin{tabular}[c]{@{}c@{}}Average \\ Results\end{tabular}}&\multirow{2}{*}{\begin{tabular}[c]{@{}c@{}}Standard \\ Deviation\end{tabular}}& \multicolumn{5}{c}{Testing subjects (Target Domains)} \\
\cline{5-9} 
& &&& $N_{1}$ & $N_{2}$ & $N_{3}$ & $N_{4}$ & $N_{5}$\\ \midrule
\multirow{3}{*}{4} &  MAE (mg/dL)&$23.28$ &$9.08$& $19.10$ & $27.25$ & $19.33$ & $37.03$ &$13.69$ \\
                   & RMSE (mg/dL) &$40.88$&$15.99$& $36.42$ & $46.25$ & $27.32$ & $66.05$ &$28.38$\\
                   & MAPE (\%) &$14.10$&$4.20$ &$17.22$ & $17.87$ & $13.63$ & $14.47$ &$7.31$\\
\cmidrule{1-9}
\multirow{3}{*}{5} &  MAE (mg/dL) &$20.95$&$9.59$& $16.53$ & $26.40$ & $16.91$ & $34.67$ &$10.26$\\
                   & RMSE (mg/dL) &$37.28$&$16.84$& $27.27$ & $47.64$ & $25.23$ & $62.00$ &$24.24$\\
                   & MAPE (\%) &$12.40$&$4.31$& $13.54$ & $17.00$ & $12.22$ & $13.90$ &$5.35$\\
\cmidrule{1-9}
\multirow{3}{*}{6} & MAE (mg/dL) &$30.95$&$17.59$& $22.48$ & $31.76$ & $23.31$ & $60.86$&$16.36$ \\
                   & RMSE (mg/dL) &$51.72$&$26.56$& $42.93$ & $53.53$ & $31.15$ & $96.63$ &$34.34$\\
                   & MAPE (\%) &$18.40$&$6.59$& $17.98$ & $20.72$ & $16.76$ & $27.37$ &$9.18$\\
\cmidrule{1-9}
\multirow{3}{*}{7} &  MAE (mg/dL) &$19.80$&$6.14$& $17.72$ & $23.89$ & $17.01$ & $27.99$ & $12.38$ \\
                   & RMSE (mg/dL) &$34.40$&$9.48$& $30.64$ & $41.21$ & $24.43$ & $47.24$ & $28.49$ \\
                   & MAPE (\%) &$12.22$&$3.23$& $14.19$ & $14.61$ & $11.76$ & $13.75$ & $6.78$ \\
\cmidrule{1-9}
\multirow{3}{*}{8} & MAE (mg/dL) &$17.47$& $5.42$&$15.65$ & $21.22$ & $16.38$ & $24.08$ & $10.03$ \\
                   & RMSE (mg/dL) &$31.83$& $9.89$&$26.57$ & $41.45$ & $24.93$ & $43.59$ & $22.59$ \\
                   & MAPE (\%) &$10.88$&$3.22$ &$10.55$ & $14.31$ & $11.04$ & $12.74$ & $5.76$ \\
\cmidrule{1-9}
\multirow{3}{*}{9} & MAE (mg/dL) &$26.47$&$8.31$& $22.75$ & $28.34$ & $22.94$ & $39.92$ & $18.39$ \\
                   & RMSE (mg/dL) &$45.69$&$14.59$ &$43.41$ & $47.84$ & $32.73$ & $69.37$ & $35.08$ \\
                   & MAPE (\%) &$16.00$&$3.33$ &$17.16$ & $18.01$ & $16.08$ & $18.49$ & $10.28$ \\
\bottomrule
\end{tabular}
\label{tab:my-table_3}
\end{table*}

{\color{black}For experiments 7, 8, and 9, which utilize the Meta-forests model, a meta-learning approach is integrated with the leave-one-domain-out strategy. Similar to the Random Forests model, Meta-forests first divide all domains into target and source domains. However, unlike utilizing all source domains in one training round, they are subdivided into meta-train and meta-test sets as per the meta-learning approach, generating Random Forests models across several meta-task iterations. At each iteration, 30\% of one subject $N_{i}$ ($P=30\%$) is set as the meta-test set, with the remaining subjects preparing the meta-train set. The weights of the Random forests generated during each meta-task iteration are updated following the Meta-forests' weight update function. The final prediction of Meta-forests is a weighted average of outputs from the Random forest models generated across several multiple iterations. Table \ref{tab:my-table_4} presents an ablation analysis of different portions $P$ used in each meta-task iteration, with $P=30\%$ identified as the optimal portion of the meta-train set for each iteration.

The results in Table \ref{tab:my-table_3} provide an evaluation of the personalized model experiments, illustrating the performance of both Random Forests and Meta-forests models under the leave-one-domain-out strategy. The table presents the results when each subject $N_{i}$ served as the testing set. The 'average results' column displays the mean of three evaluation metrics across five different subjects used as the testing set. Notably, the results reveal variability in predictive accuracy across different subjects $N_{i}$ ($N_{i}, i = 1, 2, ..., 5$), as indicated by MAE, RMSE, and MAPE metrics. This variation emphasizes that model performance is subject-specific, with certain subjects achieving higher prediction accuracy than others. This variability can be attributed to the physiological differences among subjects, which affect the blood glucose interaction with mm-wave at specific frequencies.

Despite these individual differences, a consistent trend of improved accuracy with selected mm-wave frequencies is observed, suggesting that personalized models can be enhanced through mm-wave frequency selection. Notably, the experiments utilizing the Meta-forests model (Experiments 7, 8, and 9) generally show improved performance compared to those using the Random Forests model (Experiments 4, 5, and 6). This improvement suggests that the Meta-forests model, with its integration of meta-learning strategies, is adept at handling individual variability and domain-specific features. Furthermore, the standard deviation values in Table \ref{tab:my-table_3} emphasize the Meta-forests' capability to find a generalized sample space across various different distribution domains. }

\section{Discussion}
\subsection{Comparison with Other Related Glucose Monitoring Models}

As presented in the Introduction section, several related works have developed non-invasive glucose prediction models using their own collected human-test glucose datasets. According to our investigation, these studies primarily focus on generalized experiments rather than the personalized experiments we have conducted. Table \ref{tab:add_1} provides a comparison of the results from generalized experiments, including those from related works and our application of random forests. While our model's prediction results are competitive, they do not exhibit the lowest error in generalized experiments. We attribute this to differences in glucose measurement characteristics and collection methods.

% \begin{table}[htb!]
% \centering
% \caption{The Generalized Experiments Comparison Results with Related Glucose Monitoring Models. Some entries are noted as 'N/A', denoting that specific metrics were not evaluated in the original related study.}
% \label{tab:add_1}
% \begin{tabular}{lccc}
% \toprule
% \multirow{2}{*}{Models}& \multirow{2}{*}{Sensors}  &  RMSE  &  MAPE \\
% & & (mg/dL) &  (\%) \\
% \midrule
% Lee et al.\cite{cite_intro05} & PPG  &31.90 &N/A \\
% Zhang et al.\cite{cite_intro06} & PPG       & 23.92& N/A  \\
% Zhang et al.\cite{cite_intro07} & PPG     &  35.59&  N/A  \\
% Segman et al.\cite{cite_readd01} & NIR  &N/A & 17.90\\
% Li et al.\cite{cite_readd02} &  ECG      &   37.17 &  N/A  \\

% \midrule
% \textit{\textbf{ours.}}  & Mm-wave, NIR    & 34.01 &  12.59\\ 
% \bottomrule
% \end{tabular}
% \end{table}

\begin{table}[htb!]
\centering
\caption{The Generalized Experiment Comparison Results with Related Glucose Monitoring Models. Entries marked as 'N/A' indicate that specific metrics were not evaluated in the original studies.}
\label{tab:add_1}
\begin{tabular}{lccc}
\toprule
\multirow{2}{*}{Models}& \multirow{2}{*}{Sensors}  & {RMSE}  &  {MAPE} \\
& &(mg/dL) &  (\%) \\
\midrule
{Lee et al.\cite{cite_intro05}} & {PPG}  & {31.90} & {N/A} \\
{Zhang et al.\cite{cite_intro06}} &{PPG} & {23.92} & {N/A}  \\
{Zhang et al.\cite{cite_intro07}} &{PPG} & {35.59} & {N/A}  \\
{Segman et al.\cite{cite_readd01}} &{NIR} &{N/A} & {17.90} \\
{Li et al.\cite{cite_readd02}} &{ECG}  &{37.17} & {N/A}  \\
\midrule
{\textit{\textbf{ours.}}}  & {Mm-wave, NIR} & {34.01} & {12.59} \\ 
\bottomrule
\end{tabular}
\end{table}

In our study, we introduced a novel approach by conducting personalized experiments that use data from different individuals for model training and testing. These experiments are designed to address the realistic challenge of not having a sufficiently large dataset to account for variability among subjects, which is a factor largely overlooked in related works. Due to the inherent heterogeneity among subjects, glucose metabolism varies significantly. Some individuals experience a rapid drop in glucose levels post-injection, while others do not. Consequently, as shown in Table \ref{tab:add_2}, our collected dataset exhibits the largest range and standard deviation of glucose readings compared to related works, leading to higher prediction errors. The relatively high standard deviations and evaluation metrics in Table \ref{tab:my-table_3} further support this observation. However, our applied Meta-forests model significantly reduces these standard deviations and evaluation metrics when comparing Experiments 7, 8, and 9 to 4, 5, and 6. This demonstrates the model's generalizability to unseen subjects and its potential for clinical deployment.

Additionally, the methods of glucose reading collection differ between these works and ours, which results in variations in glucose levels. Our glucose readings were collected through an intravenous glucose tolerance test (IVGTT), where subjects fasted for 12 hours before receiving an intravenous glucose injection. Conversely, the dataset collected by Li et al. \cite{cite_readd02} utilized an oral glucose tolerance test (OGTT), where subjects ingested glucose orally after prolonged fasting. Both IVGTT and OGTT result in rapid fluctuations in blood glucose levels over a short period. Consequently, the variations and standard deviations of glucose readings are large, leading to increased prediction errors as shown in Table \ref{tab:add_1}. In contrast, the first four glucose reading datasets listed in Table \ref{tab:add_2} were collected via continuous monitoring (CM), which involves routine clinical measurement of glucose levels without additional experimental interventions. Due to glucose metabolic homeostasis, glucose readings from CM remain relatively stable, resulting in lower prediction errors for the corresponding models in Table \ref{tab:add_2}.

% \begin{table}[htb!]
% \centering
% \caption{The Characteristics of the Glucose Readings of Comparison Models.}
% \label{tab:add_2}
% \begin{tabular}{lcc}
% \toprule
% \multirow{2}{*}{Models}& Collection& (Min, Max) $\pm$ Standard deviation \\
% &Methods&(mg/dL) \\
% \midrule
% Lee et al.\cite{cite_intro05} & CM &  (40, 270) \\
% Zhang et al.\cite{cite_intro06}  & CM    & (75.6, 183.6) $\pm$ 27.0\\
% Zhang et al.\cite{cite_intro07}  & CM   & (70.2, 198.0) $\pm$ 38.9 \\
% Segman et al.\cite{cite_readd01}& CM&  (69.0, 402.0) $\pm$ 70.0\\
% Li et al.\cite{cite_readd02}   & OGTT  &   N/A  \\
% \midrule
% \textit{\textbf{ours.}} & IVGTT  & (37.8, 547.2) $\pm$ 107.0 \\ 
% \bottomrule
% \end{tabular}
% \end{table}

\begin{table}[htb!]
\centering
\caption{The Characteristics of Glucose Readings from Various Comparison Models}
\label{tab:add_2}
\begin{tabular}{lcc}
\toprule
\multirow{2}{*}{Models}& {Collection}& {(Min, Max) $\pm$ Standard deviation} \\
&{Methods}&{(mg/dL)} \\
\midrule
{Lee et al.\cite{cite_intro05}} & {CM} & {(40, 270)} \\
{Zhang et al.\cite{cite_intro06}}  & {CM}    & {(75.6, 183.6) $\pm$ 27.0} \\
{Zhang et al.\cite{cite_intro07}}  & {CM}   & {(70.2, 198.0) $\pm$ 38.9} \\
{Segman et al.\cite{cite_readd01}} & {CM} &  {(69.0, 402.0) $\pm$ 70.0} \\
{Li et al.\cite{cite_readd02}}   & {OGTT}  &  {N/A}  \\
\midrule
{\textit{\textbf{ours.}}} & {IVGTT}  & {(37.8, 547.2) $\pm$ 107.0} \\ 
\bottomrule
\end{tabular}
\end{table}

\subsection{Correlation and Mixed Model Analysis}

\begin{figure}[htb!]
    \centering
    \includegraphics[width = 1\linewidth,height = 0.9\linewidth]{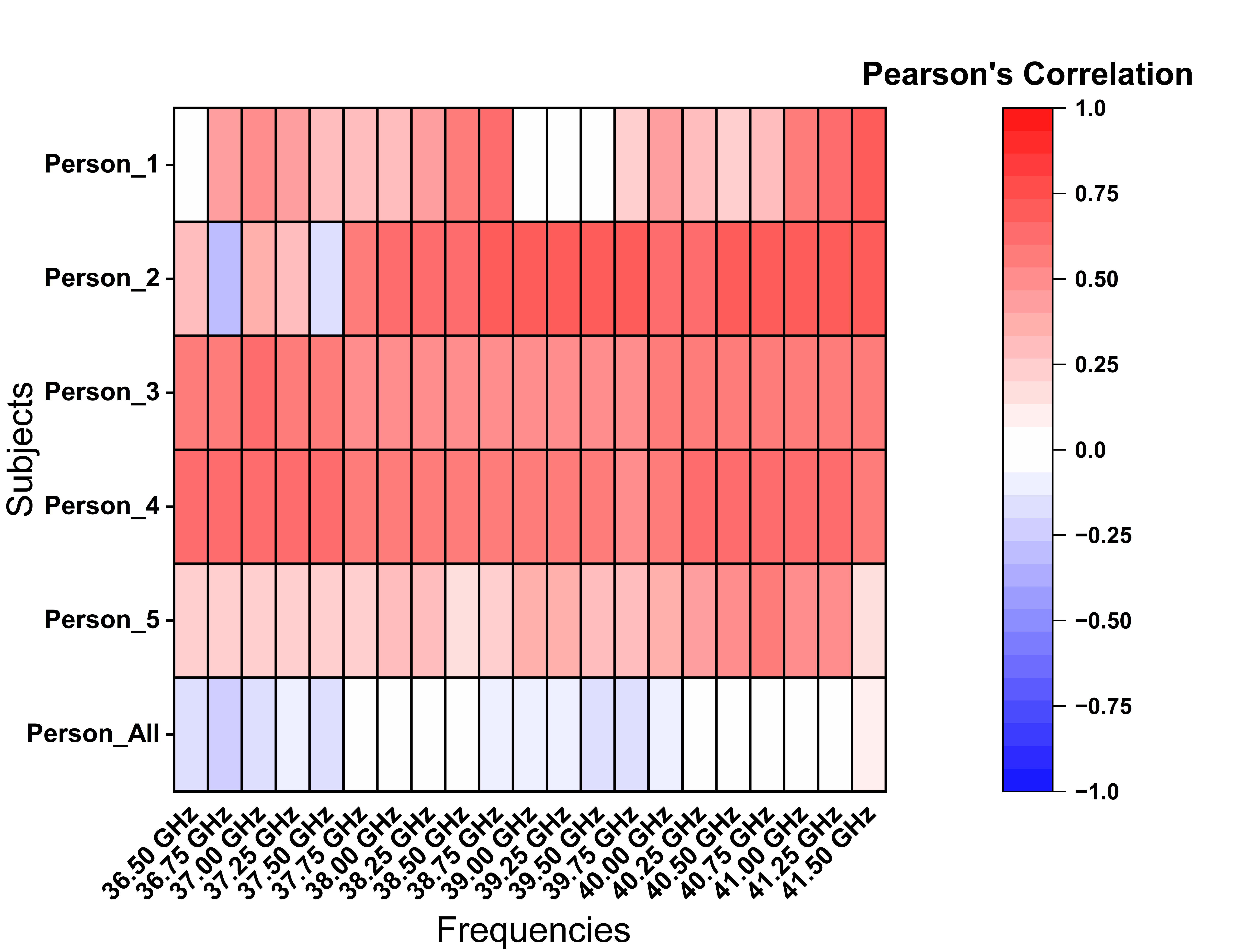}
    \caption{Pearson's Correlation Coefficients Between $S_{21}$ Parameters and Blood Glucose Values Across $N_{i}$ ($N_{i}, i = 1, 2, ..., 5$) Subjects.}
    \label{fig:Heatmap}
\end{figure}

Our previous study with these datasets featured an analysis using Pearson's correlation coefficients \cite{cite_D1} to assess the linear relationships between the $S_{21}$ parameters of mm-wave frequencies and blood glucose values. Ranging from -1 to 1, this measure is invaluable for its simplicity in identifying direct linear associations between two continuous variables. However, it is limited in datasets with heterogeneity among subjects. This limitation was particularly evident in Pearson's correlation heatmap (refer to Figure \ref{fig:Heatmap}), which displayed varying correlation strengths across different subjects. It suggested that some frequencies significantly correlate with glucose values in certain individuals but not in others. Notably, as indicated by the last row of the heatmap, considering all subjects' data collectively without identifying individual variance could potentially lead to misleading results. This variability suggested the presence of heterogeneity among subjects affecting the relationship between $S_{21}$ parameters and glucose levels, which Pearson's correlation could not address.

To address these challenges, we implemented a MixedLM. This statistical tool, comprising both fixed and random effects, enabled us to analyze associations between features and labels while considering heterogeneity among subjects. In our dataset, the fixed effects were represented by the $S_{21}$ parameters of mm-wave, which were hypothesized to influence blood glucose values. In contrast, random effects accounted for individual variability among subjects, a crucial aspect considering the personalized nature of glucose metabolism.  This model allowed us to isolate the effect of each mm-wave frequency on glucose values while adjusting for inter-subject differences.

\subsection{Subject Heterogeneity Analysis}

Heterogeneities among the same subject cannot be neglected when developing a ubiquitous prediction model with low errors across different subjects. The heterogeneity in glucose monitoring can be categorized into two main types: behavioral and physiological.

Behavioral heterogeneity involves differences in lifestyle choices and behaviors among subjects, including dietary habits, physical activity levels, and sleep patterns. To minimize the effects of behavioral heterogeneity, the IVGTT data used in our study was collected over comparatively shorter periods during data collection, ensuring more stable conditions compared to other related works.

Physiological heterogeneity includes metabolic differences and variations in biomedical and anthropometric characteristics among individuals. These inherent differences can lead to variability in baseline glucose levels and glucose responses to insulin injections or carbohydrate intake. Therefore, we introduced a domain generalization method, Meta-forests, to address the dataset distribution variance caused by these inherent differences. The results in Table \ref{tab:my-table_3} demonstrate the proposed model’s ability to generalize its performance to unseen subjects.

\subsection{Limitations and Future Work}

The inherent penetration depth limits of NIR signals pose challenges for glucose detection. Compared to related spectroscopic technologies such as Raman and mid-infrared spectroscopy, NIR signals have superior penetration through human tissues, typically exceeding 0.5 mm \cite{rabinovitch1982noninvasive}. However, transmitting NIR signals into deeper human tissue to capture glucose information remains difficult. To mitigate this limitation, the data used in this study was collected from the thin tissue between the thumb and index finger to reduce the penetration limitation of NIR signals. Additionally, mm-wave signals have significantly higher penetration depths in human tissue. Thus, mm-wave sensors were integrated into the sensing system to detect deeper tissue information. The distinct detection mechanisms of NIR and mm-wave signals after entering the human body allow us to capture different types of information. The transmittance features of NIR signals are characterized by the light absorption of various molecules in the medium, whereas the S parameters of mm-wave signals detect the dielectric properties of the medium. Findings from our previous work, along with the experimental results presented, demonstrate the feasibility of integrating NIR and mm-wave sensors to capture diverse information for glucose predictions.

Moreover, as shown in Fig \ref{fig:MixedLM}, the lack of robust statistical associations between the $S_{21}$ parameters of certain frequencies and glucose changes primarily reflects differences in measurement sensitivity. Factors such as interference, antenna performance, and noise levels play a critical role in the precision of measurements at specific frequencies, thereby potentially affecting the detection of glucose level changes. However, this postulation requires further experimental validation, to test our data-driven approach of uncovering mechanisms. As preliminary work, our findings lay the groundwork for future research aimed at exploring the complex associations between sensing features and biological parameters. Subsequent studies could focus on exploring these findings within larger datasets to both validate our hypotheses and enhance the knowledge of non-invasive glucose detection technologies.

We note that employing a repetitive construction loop in the modified Meta-forests model results in additional training time compared to the basic Random Forests model. Currently, the Meta-forests model maintains relatively low computational times given the limited data size. However, as the dataset grows, the training time is expected to increase significantly. To address this, we plan to implement optimization techniques such as parallel processing in future work to enhance computational efficiency.

This study serves as a pilot study that includes a limited sample, providing a feasibility validation of the established glucose prediction system. To ensure the generalizability and reliability of the prediction system, future studies will incorporate a broader range of participants. This expansion will include individuals from varying age groups, ethnicities, and health conditions to ensure that our system performs reliably across diverse users. We plan to undertake multi-center trials to evaluate the device under different environmental and clinical settings, thereby enhancing the robustness of the prediction system and improving the device's predictive accuracy in real-world scenarios.

Compliance with regulatory standards for medical devices is of critical importance. This study achieved a MAPE of 10.88\%, which is competitive with existing glucose meters as documented in \cite{heinemann2020benefits}. However, progressing to full regulatory certification, such as approval from the Food and Drug Administration, involves complex challenges. These challenges include conducting rigorous testing with a larger number of participants, which simultaneously increases the difficulty of ensuring their privacy. Beyond the non-disclosure agreements and ethical permissions previously mentioned, we are exploring the feasibility of incorporating federated learning models in future studies, which would allow for data analysis without centralizing sensitive information, thus significantly enhancing privacy protections and reducing the risk of data breaches.

\section{Conclusion}
This study contributes to non-invasive glucose detection by integrating NIR spectroscopy and mm-wave sensing in a multi-sensor system. Our approach, employing a MixedLM for feature analysis, demonstrated promising results in blood glucose prediction, showing potential for clinical application. A significant aspect of our work involves the application of the DG model, Meta-forests, showcasing its ability to effectively handle domain variance in the dataset. This model's incorporation into our predictive system highlights its adaptability to individual variances, further enhancing the accuracy and reliability of non-invasive glucose detection.

\end{document}